%% file: paper_camera_ready.tex
\documentclass[10pt,twocolumn,letterpaper]{article}

\usepackage{wacv}              

\usepackage{graphicx}
\usepackage{amsmath}
\usepackage{amssymb}
\usepackage{amsfonts}
\usepackage{booktabs}
\usepackage{calc}
\usepackage{adjustbox}
\usepackage{algorithm2e}
\usepackage{pifont}
\usepackage{float}
\usepackage[misc]{ifsym}

\def\usepng{0}  

\newlength\myheight
\newlength\mydepth
\settototalheight\myheight{Xygp}
\usepackage[pagebackref,breaklinks,colorlinks]{hyperref}

\usepackage[capitalize]{cleveref}
\crefname{section}{Sec.}{Secs.}
\Crefname{section}{Section}{Sections}
\Crefname{table}{Table}{Tables}
\crefname{table}{Tab.}{Tabs.}

\def\wacvPaperID{552} 
\def\confName{WACV}
\def\confYear{2025}

\begin{document}

\title{GAUDA: \textit{Generative Adaptive Uncertainty-guided Diffusion-based Augmentation} for Surgical Segmentation}

\author{
Yannik Frisch$^{1,2}$ \textsuperscript{(\Letter)}, Christina Bornberg$^3$, Moritz Fuchs$^1$, Anirban Mukhopadhyay$^1$\\
$^1$ Technical University Darmstadt, Fraunhoferstr. 5, 64285 Darmstadt, GER\\
$^2$ University Medical Center Mainz, Langenbeckstr. 1, 55131 Mainz, GER\\
$^3$ University of Girona, Placa de Sant Domenec 3, 17004 Girona, ESP\\
{\tt\small yannik.frisch@gris.tu-darmstadt.de}
}

\maketitle

\begin{abstract}
   Augmentation by generative modelling yields a promising alternative to the accumulation of surgical data, where ethical, organisational and regulatory aspects must be considered. Yet, the joint synthesis of (image, mask) pairs for segmentation, a major application in surgery, is rather unexplored. We propose to learn semantically comprehensive yet compact latent representations of the (image, mask) space, which we jointly model with a Latent Diffusion Model. We show that our approach can effectively synthesise unseen high-quality paired segmentation data of remarkable semantic coherence. Generative augmentation is typically applied pre-training by synthesising a fixed number of additional training samples to improve downstream task models. To enhance this approach, we further propose Generative Adaptive Uncertainty-guided Diffusion-based Augmentation (GAUDA), leveraging the epistemic uncertainty of a Bayesian downstream model for targeted online synthesis. We condition the generative model on classes with high estimated uncertainty during training to produce additional unseen samples for these classes. By adaptively utilising the generative model online, we can minimise the number of additional training samples and centre them around the currently most uncertain parts of the data distribution. GAUDA effectively improves downstream segmentation results over comparable methods by an average absolute IoU of 1.6\% on CaDISv2 and 1.5\% on CholecSeg8k, two prominent surgical datasets for semantic segmentation.
\end{abstract}

\input{sections_camera_ready/01_intro}
\input{sections_camera_ready/02_background}
\input{sections_camera_ready/03_method}
\input{sections_camera_ready/04_exps}
\input{sections_camera_ready/05_results}
\input{sections_camera_ready/06_discussion}
\input{sections_camera_ready/07_conclusions}
\clearpage

{\small
\bibliographystyle{ieee_fullname}
\bibliography{egbib}
}

\clearpage
\input{sections_camera_ready/supplementary}

\end{document}

%% file: sections_camera_ready/01_intro.tex
\section{Introduction}
Deep Learning (DL) models are crucial for surgical assistance systems. However, these models often fail in generalising to rare cases~\cite{sahu2021simulation}, which often occur due to the sequential nature, anatomical variability and functional requirements of surgical datasets. While there have been significant achievements in DL-based segmentation of surgical recordings, model performance heavily depends on the amount and variety of training data \cite{frisch2023synthesising,hong2020cholecseg8k,luengo20212020}.

\begin{figure*}[htbp]
    \centering
    \if\usepng1
        \includegraphics[width=\textwidth]{figures/infographics_v4.png}
    \else
        \includegraphics[width=\textwidth]{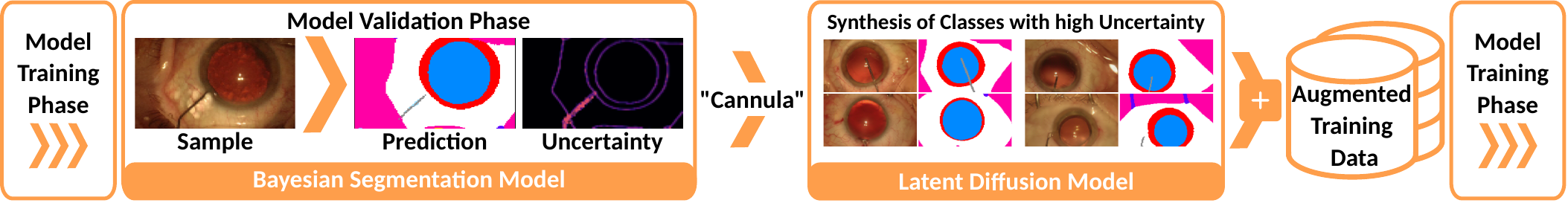}
    \fi
    \caption{\textbf{Abstract overview of GAUDA.} Given the class-wise epistemic uncertainty of a \textit{Bayesian Segmentation Model}, we adaptively synthesise new samples during the validation phase utilising a \textit{Latent Diffusion Model} conditioned on those labels, shown here for the \textit{Cannula} tool. Incorporating the synthetic data into the model's training data boosts performance for uncertain cases.} 
    \label{fig:introfig}
\end{figure*}

\textbf{Accumulating additional surgical training data} involves collecting and annotating new data -- an arduous procedure, especially for \textbf{pixel-based segmentation} -- and is potentially hindered by regulatory and ethical aspects \cite{lennerz2024dimensions}. Alternatively, researchers resort to traditional data augmentation techniques \cite{shorten2019survey}. However, these methods often fail to escape the confines of the original data's visual semantics. To alleviate this, recent advancements included \textbf{generative approaches}, like GANs or Denoising Diffusion Models (DDMs), to synthesise novel training data across various domains \cite{antoniou2017data,chen2021synthetic,chlap2021review,frisch2023synthesising,kazerouni2023diffusion,sagers2023augmenting,sandfort2019data,shin2018medical,trabucco2023effective,wang2021towards}. 

Generative augmentations typically synthesise a fixed number of additional training samples, which might not contribute to a model's progress. We propose an alternative direction by \textbf{guiding the generation in an online fashion} - essentially extending Adaptive Sampling \cite{gopal2016adaptive} to generative adaptive augmentation. Instead of relying on rather uninformative validation scores, we propose employing Bayesian principles \cite{baumgartner2019phiseg,fuchs2021practical,kohl2018probabilistic} to estimate a downstream model's epistemic uncertainty. This insight allows our pipeline, as illustrated in Figure \ref{fig:introfig}, to \textbf{adaptively generate new samples for classes with higher uncertainty}. Such guided generative augmentation efficiently and dynamically identifies the specific deficiencies in the training dataset.

For modelling the joint (image, mask) distribution of segmentation data, Korneliusson et al. \cite{korneliusson2019generative} propose a progressively growing conditional Style-GAN \cite{karras2020analyzing} to generate images of people along with segmentation masks for clothing. DatasetGAN \cite{zhang2021datasetgan} trains a decoder on StyleGAN feature maps to produce segmentation masks from a minimal number of paired examples. SinGAN-Seg \cite{thambawita2022singan} builds upon style transfer to generate endoscopy image data along with binary segmentation masks for polyps. Park et al. \cite{park2023learning} propose a Categorical Diffusion Process to model the joint distribution in image space and train an additional super-resolution model to produce high-quality images. We propose an orthogonal and more \textbf{resource-efficient direction for modelling paired surgical data} by learning separate semantically compressing representations of the images and masks utilising VQ-GAN \cite{esser2021taming}. We then model the joint distribution of these representations using a Latent Diffusion Model (LDM) \cite{rombach2022high}, which allows for robust and resource-efficient synthesis of paired data and can be viewed as a multi-modal LDM \cite{bao2023one,huang2023collaborative,muller2023multimodal,nair2023unite,ruan2023mm}.

Through our novel \textbf{online integration of generative augmentation for segmentation data}, we expand the usage of DL-based analysis and assistance methods for surgery. We introduce a new perspective on how generative augmentation for Surgical Data Science is leveraged more sensibly and tailored to the model's evolving needs.
Our approach significantly boosts the model's performance for surgical segmentation by \textbf{effectively and adaptively expanding the training dataset where it is most needed}.

\paragraph{Contributions:}
\begin{enumerate}
    \item We are the first to propose a resource-efficient method based on Latent Diffusion Models to synthesise high-quality and semantically coherent paired (image, mask) data for surgical segmentation.
    \item We further propose GAUDA, a novel adaptive and generative augmentation scheme guided by the epistemic uncertainty of a downstream task model, yielding a greater performance increase than comparable methods.
\end{enumerate}

%% file: sections_camera_ready/02_background.tex
\section{Background}
\label{chap:bg}

Generative models approximate a data domain's distribution $p(x)$. The idea behind DDMs is to sequentially diffuse image samples $x_0 \sim p(x_0)$ from the unknown data distribution using univariate Gaussian noise. Eventually, after $T$ steps, the chain $x_1, ..., x_T$ reaches a representation close to pure Gaussian noise for $T \rightarrow \infty$. Assuming adequate step sizes, a sufficiently large $T$, and a well-behaved variance schedule $\{\beta_1, ..., \beta_T\} \in (0,1)^T$, one can train a parameterised denoising model $\epsilon_\theta$ to revert individual steps. Since $p(x_T)$ is a known distribution, we can sample $x_T \sim \mathcal{N}(0,\mathbf{I})$ from it and iteratively apply the denoising model to push it back into the unknown data distribution. As shown by Ho et al. \cite{ho2020denoising}, the training of the denoising model can be reduced to the simplified loss term 
\begin{multline}
    \label{eq:L_simple}
    \mathcal{L}_\text{simple}(\theta) := \mathbb{E}_{x_0,\epsilon,t}[ ||\epsilon - \epsilon_\theta(\sqrt{\Bar{\alpha}_t}x_0 + \sqrt{1 - \Bar{\alpha}_t}\epsilon,t)||_2^2]
\end{multline}
with $\epsilon \sim \mathcal{N}(\mathbf{0}, \mathbf{I})$ and the factorisations $\alpha_t := 1 - \beta_t$ and $\Bar{\alpha}_t := \prod_{s=1}^t \alpha_s$.

The efficiency of DDMs was significantly improved due to the introduction of LDMs \cite{rombach2022high}. They compress the high-dimensional $x$ into a lower-dimensional latent representation $z$ using a pre-trained autoencoder architecture, e.g. a variant of VQ-GAN \cite{esser2021taming}. These architectures consist of an encoding part $z_0 = \mathcal{E}(x_0)$ and a decoder $\Tilde{x}_0 = \mathcal{D}(\mathcal{E}(x_0))$. Using these pre-trained architectures, the diffusion is performed in the latent space, with the denoising model $\epsilon_\theta$ optimised via
\begin{equation}
    \label{eq:L_LDM}
    \mathcal{L}_\text{LDM}(\theta) = \mathbb{E}_{z_t,\epsilon,t}[||\epsilon - \epsilon_\theta(z_t,t)||^2_2]
\end{equation}
where $z_t$ can be obtained as $z_t = \sqrt{\Bar{\alpha}_t}z_0 + \sqrt{1 - \Bar{\alpha}_t}\epsilon$.

\subsection{Adaptive Sampling}

Usually, class imbalances in downstream tasks are combated by pre-computing sampling weights for data points \cite{sahu2017addressing}. For example, the weights $w_c$ of class $c$ can be defined as $w_c = 1/\sqrt{f(c)}$ where $f(c)$ is the frequency of the label appearing in the training dataset. This choice can result in a suboptimal selection of training samples. As an alternative, Gopal et al. \cite{gopal2016adaptive} proposed adapting the weights online during training, based on validation scores. 

Adaptive Sampling (AS) offers a dynamic and responsive approach to addressing class imbalance, ensuring that the sampling process evolves with the model's learning state. We further enhance this approach by generative augmentation and using a model's predictive uncertainty to redefine sampling weights. An analysis of this modelling choice based on a simple simulation can be found in Supplementary Section \ref{supp:sampling}, showing that uncertainty-based sampling can yield faster convergence and better generalisation over score-based sampling.

%% file: sections_camera_ready/03_method.tex
\section{Method}
\label{chap:method}

\begin{figure*}[htbp]
    \centering
    \if\usepng1
        \includegraphics[width=\textwidth]{figures/method_figure_v4.png}
    \else
        \includegraphics[width=\textwidth]{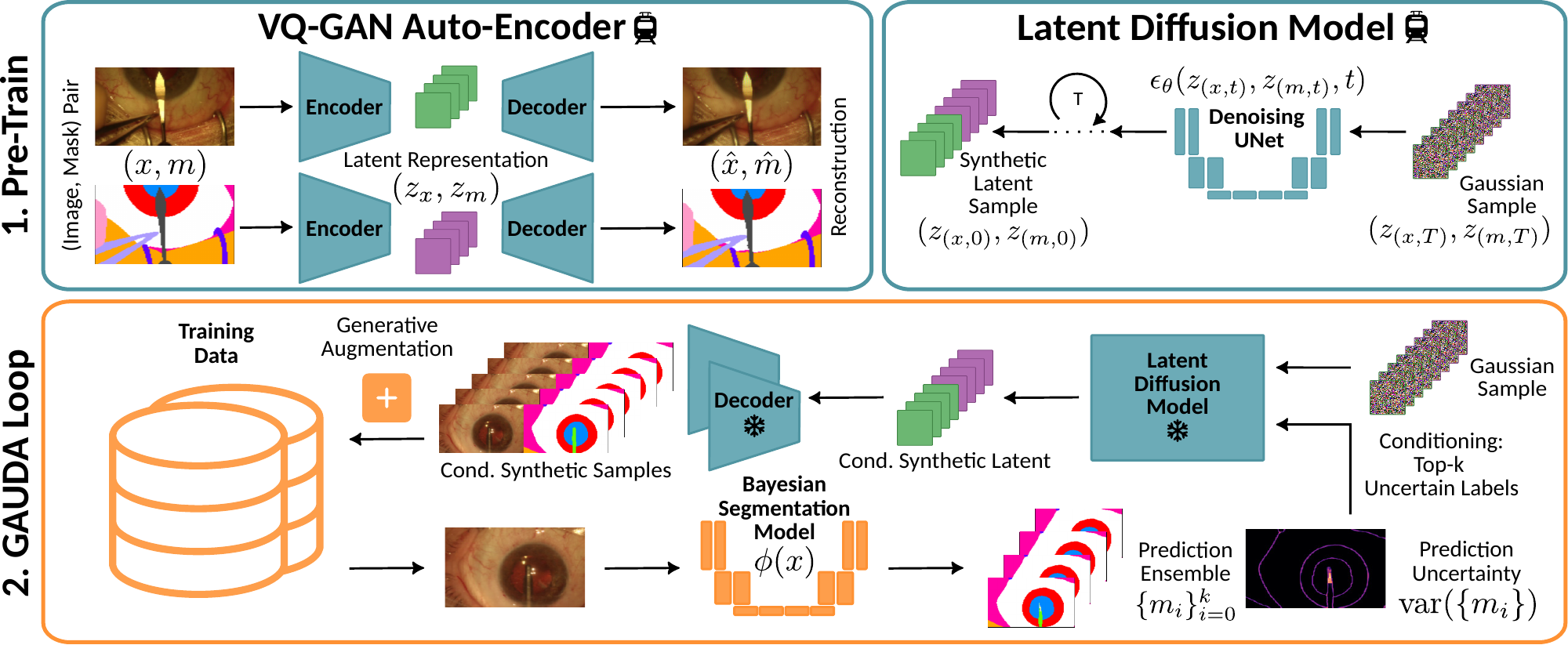}
    \fi
    \caption{\textbf{Semantic latent diffusion architecture and GAUDA pipeline.} Using pre-trained (train) image and mask VQ-GANs (top left), we obtain a combined latent space, which we approximate with a pre-trained (train) LDM (top right). GAUDA leverages the frozen (snowflake) LDM in an adaptive online fashion by synthesising additional training samples conditioned on labels with high predictive uncertainty (bottom).} 
    \label{fig:methodfig}
\end{figure*}

We start by explaining our generative model for synthetic (mask, image) pairs for segmentation data. We then briefly describe the downstream task model for segmentation before providing an overview of GAUDA, shown in Figure \ref{fig:methodfig}.

\subsection{Generative Model for Segmentation Data}

Training DDMs for directly \textbf{modelling the joint distribution $p(x,m)$ of (image, mask) pairs} would significantly increase the required compute. Therefore, we follow the LDM formulation \cite{rombach2022high} to instead approximate the distribution of latent space representations $p(z)$, preceded by two autoencoders, representing $p(z|x,m)$ and $p(x,m|z)$.

\begin{figure}[htbp]
    \centering
    \if\usepng1
        \includegraphics[width=\linewidth]{figures/param_flops_plot_v3.png}
    \else
        \includegraphics[width=\linewidth]{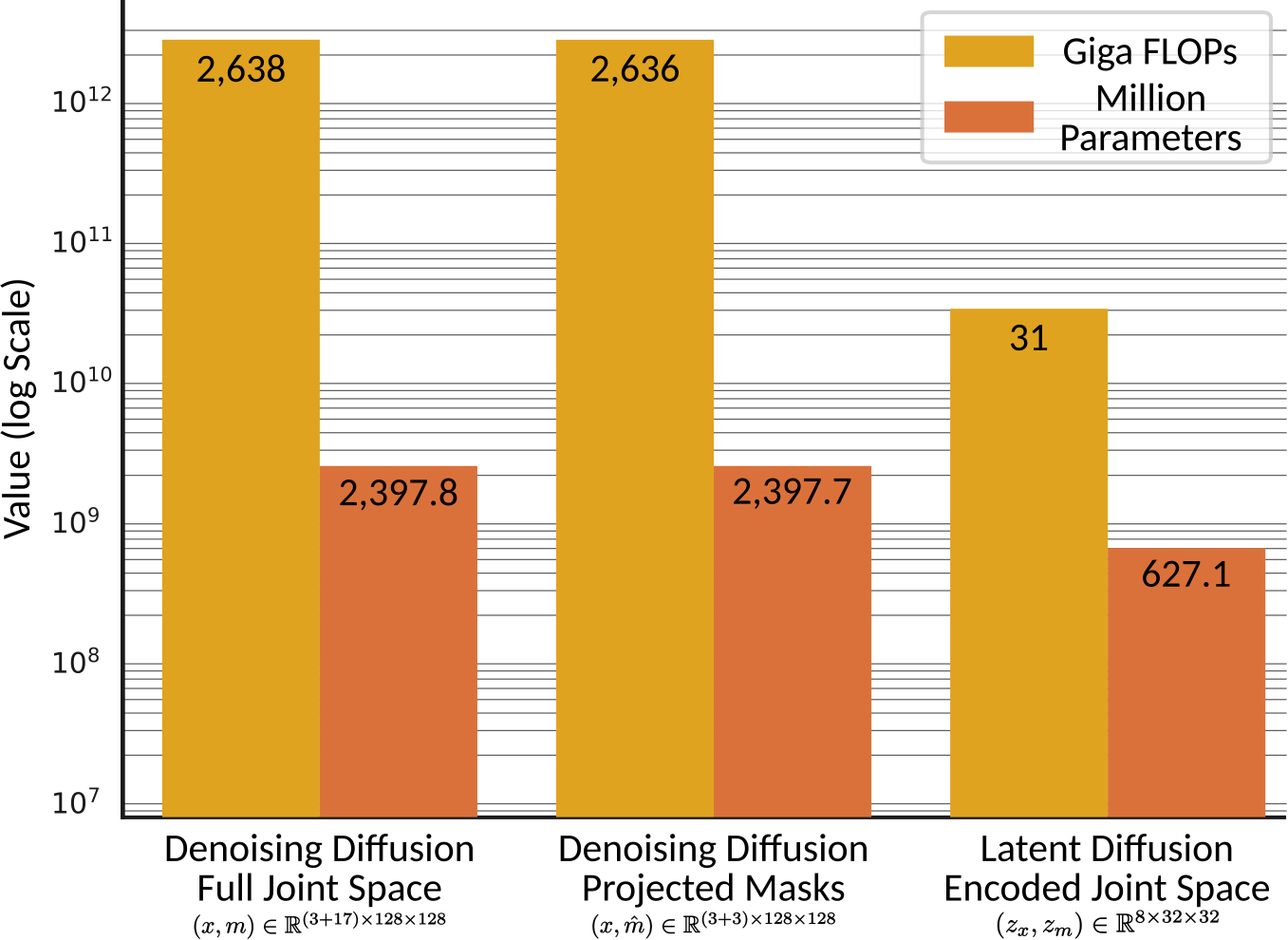}
    \fi
    \caption{\textbf{Denoising model FLOPS and parameter count.} Jointly modelling (image, mask) pairs in an auto-encoded latent space allows for more efficient denoising in the encoded space, significantly reducing the required number of floating point operations and parameters.} 
    \label{fig:paramflops}
\end{figure}

Our autoencoders consist of separate VQ-GANs \cite{esser2021taming} which are trained to map images $x \in \mathbb{R}^{3 \times H \times W}$ and masks $m \in \mathbb{R}^{K \times H \times W}$ into latent space representations $(z_x,z_m) \in \mathbb{R}^{2d \times h \times w}$. Since $h < H, w < W$ and $2d < (K+3)$, we are effectively reducing the generative model's training space dimensions by a large margin. This reduction significantly increases the resource efficiency over comparable image space generative models for segmentation data \cite{korneliusson2019generative,park2023learning,thambawita2022singan}, as visualised in Figure \ref{fig:paramflops}.

The autoencoders' decoding parts map the latent representations $(z_x,z_m)$ back into a reconstructed (image, mask) pair $(\hat{x}, \hat{m})$. We train the encoding of images and masks with separate losses to handle the different natures of both modalities, using
\begin{equation}
    \mathcal{L}_{AE_x} = \mathbb{E}_{\mathcal{D}_x(z_x)}[||\hat{x} - x||_2^2]
\end{equation}
and 
\begin{equation}
    \mathcal{L}_{AE_m} = \mathbb{E}_{\mathcal{D}_m(z_m)}[\text{CE}(\hat{m},m)]
\end{equation}
where $\text{CE}$ is the Cross Entropy loss.

The LDM is then trained to approximate the distribution of latent pairs $(z_x,z_m)$. We utilise the pre-trained encoders $\mathcal{E}_x$ and $\mathcal{E}_m$ for images and masks and optimise the parameterised denoising model $\epsilon_\theta$ by minimising the semantic loss
\begin{equation}
    \label{eq:L_SEM}
    \mathcal{L}_\text{SEM}(\theta) = \mathbb{E}_{z_x, z_m,\epsilon,t}[||\epsilon - \epsilon_\theta(z_{(x,t)},z_{(m,t)},t)||^2_2]
\end{equation}
on noised image $z_{(x,t)}$ and mask $z_{(m,t)}$ representations, which only introduces a minimal increase in compute compared to Equation \ref{eq:L_LDM}. 

We further deploy \textit{Classifier-Free Guidance} (CFG) \cite{frisch2023synthesising,ho2021classifier} for conditional generation which substitutes the predicted noise $\epsilon_\theta(z_{(x,t)},z_{(m,t)}, t, c)$ at each step as

\begin{multline} 
    \epsilon'_\theta(z_{(x,t)},z_{(m,t)}, t, c) := (1 + \omega)\epsilon_\theta(z_{(x,t)},z_{(m,t)}, t, c) \\ - \omega \epsilon_\theta(z_{(x,t)},z_{(m,t)}, t)
\end{multline}

where $c \in \mathbb{R}^D$ is the conditioning label, $\omega$ is a scaling parameter which we empirically set to $3.0$ (see Figure \ref{fig:quan}) and $\epsilon_\theta(z_{(x,t)},z_{(m,t)}, t)$ is trained by randomly dropping the conditioning with probability $0.2$. This pre-training of the auto-encoder architecture and the LDM are visualised in the top row of Figure \ref{fig:methodfig}.\\

\subsection{Segmentation Downstream Task}

For segmentation, many models leveraging Bayesian principles have been proposed \cite{baumgartner2019phiseg,fuchs2021practical,kohl2018probabilistic}. We deploy \textit{PHiSeg} \cite{baumgartner2019phiseg} for our downstream task of surgical segmentation, which \textbf{provides reliable epistemic uncertainty estimates} without performance decreases compared to \textit{nnUNet} \cite{isensee2021nnu}. As an additional baseline and to highlight the flexibility of our approach, we additionally deploy an ensemble \cite{lakshminarayanan2017simple} of 10 UNets \cite{ronneberger2015u} as another Bayesian segmentation approach. Given an RGB image $x$, the segmentation model $S$ outputs a set of $k$ posterior mask predictions $S(x) = \{m_1, m_2, ..., m_k\}$. The final segmentation mask is obtained as the average
\begin{equation}
    m_\text{final} = \frac{1}{k}\sum_{i=1}^k m_i
\end{equation}
whereas the uncertainty estimate (UE) is derived from the variance across predictions. We define the class-wise uncertainty estimate as 
\begin{equation}
    UE_c = \frac{1}{N_c} \sum_{j \in C} \text{var}(\{m_{(1,j)}, m_{(2,j)}, ..., m_{(k,j)}\})
\end{equation}
where $C$ represents the set of pixels predicted as class $c$, $N_c$ is the total number of pixels in $C$, and $m_{(i,j)}$ is the prediction of the $i$-th mask at the $j$-th pixel. \\

\subsection{GAUDA}

While training a DL model, we can expect different subsets of the training dataset to impact the model's learning progress differently \cite{gopal2016adaptive,luengo20212020}. We \textbf{leverage epistemic uncertainty to adaptively find subsets with high impact} and train the model on the most uncertain classes.
As further analysed in Supplementary Section \ref{supp:augmentation}, adaptive augmentation yields more difficult examples than an equivalent amount of unconditional pre-training augmentation.
Our GAUDA pipeline works as follows: First, we train the downstream task model as usual (\textit{PHiSeg} $\Phi$ and \textit{UNet Ensemble} $\Upsilon$ in our case for surgical semantic segmentation). Once we reach a validation step, we estimate the model's class-wise epistemic uncertainty $UE_c$ based on validation samples. Subsequently, we synthesise a batch $\mathcal{X}_\text{syn}$ of samples for the $n_c$ classes with the highest uncertainty. We empirically choose $n_c=5$, i.e. augmenting the five most uncertain labels with synthetic data. Following that, we continue training as usual but randomly replace samples from the current original batch $\{x_i\}_{i=1}^B$ with samples from the synthetic batch $\mathcal{X}_\text{syn}$. When we reach the next validation step, this process is repeated. 
A pseudocode formulation of GAUDA can be found in the Supplementary Figure \ref{fig:pseudocode}.

Note that while we apply GAUDA to semantic segmentation here, the task can be seamlessly switched to other targets of training DL models. Further, the choices of the Bayesian DL model and UE method remain flexible and can be adapted to include future sophisticated methods. 

\begin{figure*}
    \centering
    \if\usepng1
        \includegraphics[width=0.95\textwidth]{figures/qual_examples_v3.png}
    \else
        \includegraphics[width=0.95\textwidth]{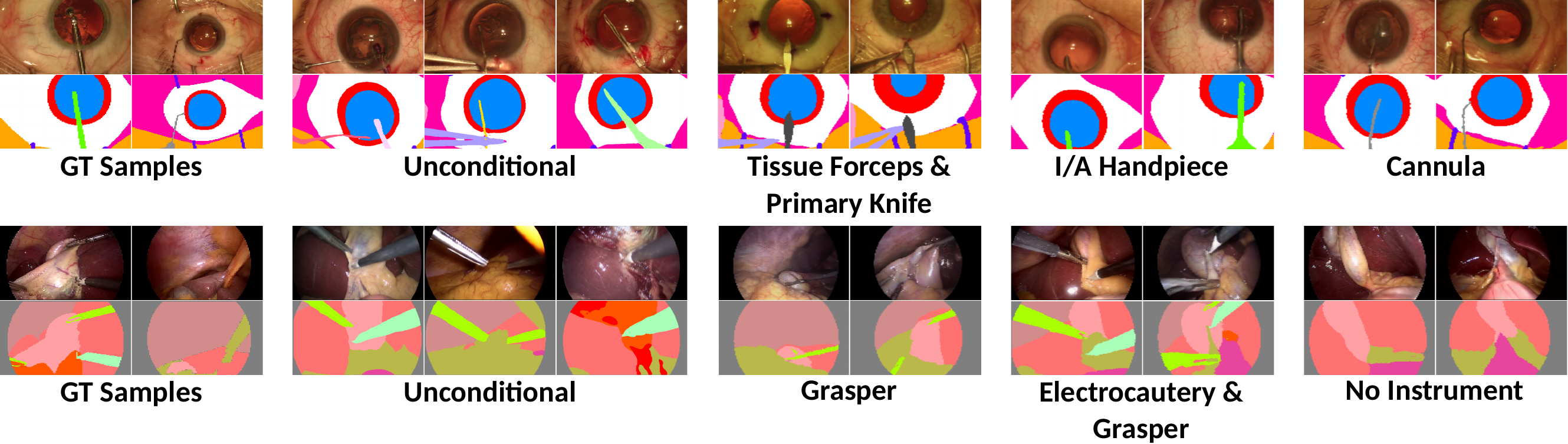}
    \fi
    \caption{\textbf{Qualitative examples.} Our generative model can synthesise high-quality and semantically aligned (image, mask) pairs for \textit{CaDISv2} Setting II (top) and \textit{CholecSeg8k} (bottom). Training with CFG allows conditional synthesis of pairs, as displayed here with tool labels in the three rightmost columns.}
    \label{fig:qual}
\end{figure*}

%% file: sections_camera_ready/04_exps.tex
\section{Datasets and Experimental Setup}

This section outlines the datasets we use for evaluating our proposed approach and gives an overview of implementation details.

\subsection{Datasets}

We consider two frequently used and distinct surgical datasets for semantic segmentation:

\textbf{CaDISv2} \cite{luengo20212020} consists of 4670 frames of cataract surgery. We randomly split the frames -- respecting the original underlying videos -- by 90\%/5\%/5\% for training, validation and testing. We use Setting II, where annotations consist of 17 labels with 10 different tools which are highly imbalanced, posing a significant difficulty to the training of DL methods.

\textbf{CholecSeg8k} \cite{hong2020cholecseg8k} contains 8080 frames of laparoscopic cholecystectomy with 13 different classes, splitting into 10 anatomy labels, 2 tools and one label for the black background. Again, we randomly split the original videos by 90\%/5\%/5\% for training, validation and testing. While containing fewer classes than CaDISv2, this dataset exhibits great variety, which stems from camera rotations and several disturbances, e.g. occlusion due to smoke or blood.

Examples of both datasets can be found in the leftmost column of Figure \ref{fig:qual}. All images were resized to $128 \times 128$ for training the VQ-GANs and the segmentation models, using bilinear interpolation. 

\subsection{Implementation Details}
This section gives an overview of the training and model details of the methods used in our experiments.

\subsubsection{Image \& Mask VQ-GAN}
We deployed VQ-GANs \cite{esser2021taming} for encoding images and masks into latent representations. They consist of encoders with 3 feature levels with [128, 256, 1024] dimensions and two residual blocks on each level. Their decoders mirror this architecture with an additional residual block on each level. Latent representations for input images of size $(3 \times 128 \times 128)$ and masks of size $(K \times 128 \times 128)$ are mapped to a shape of $(4 \times 32 \times 32)$ with a codebook size of 8192.
As a discriminator for reconstructions, we deployed the \textit{PatchGAN} discriminator from Zhu et al. \cite{CycleGAN2017} with three levels each and a feature dimension of [64, 128, 256].
We further deployed the \textit{PatchGAN} discriminator from Zhu et al. \cite{CycleGAN2017} with three levels each and a feature dimension of [64, 128, 256].
We trained the auto-encoders for 500 epochs with a batch size of 16, the \textit{AdamW} optimiser with $(\beta_1 = 0.5, \beta_2 = 0.999)$, weight decay of $1e{-5}$ and an initial learning rate of $5e{-5}$, exponentially decayed by a factor of $0.99$ after every epoch.

\subsubsection{LDM for Joint (Image, Mask) Generation}
The LDM models latent representations of the joint (image, mask) space with variables of size $(8 \times 32 \times 32)$. Its denoising UNet consists of four depth levels with [224, 448, 896, 1344] channels, respectively. Each level consists of two residual blocks. Following Dhariwal et al. \cite{dhariwal2021diffusion}, we deployed \textit{Adaptive Group Norm} for condition-aware normalisation, and \textit{Multi-Head Self-Attention} with 64 heads, applied at each resolution level ([16, 8, 4]). The model was trained for 5000 epochs with a batch size of 80. We used the \textit{AdamW} optimiser with $(\beta_1 = 0.5, \beta_2 = 0.999)$, no weight decay and an initial learning rate of $3e{-6}$, decayed polinomially with power 0.2 after each epoch. 

\subsubsection{PhiSeg for Semantic Segmentation}
We deployed \textit{PhiSeg} \cite{baumgartner2019phiseg} with 7 feature levels and [32, 64, 128, 192, 192, 192, 192] feature dimensions on each level respectively. The model was trained for 50,000 steps with a batch size of 100 using the \textit{Adam} optimiser with $(\beta_1 = 0.9, \beta_2 = 0.999)$ and a constant learning rate of $1e{-3}$. Validation was performed every 200 steps. 

\subsubsection{UNet Ensemble for Semantic Segmentation}
Our second Bayesian downstream task model consists of a deep ensemble \cite{lakshminarayanan2017simple} of $10$ UNet models \cite{ronneberger2015u}. Each member model is built with 7 feature levels and [32, 64, 128, 192, 192, 192, 192] feature dimensions on each level. The ensemble was trained with a batch size of 100 and 15,000 steps, which was sufficient for convergence. We used the \textit{Adam} optimiser with $(\beta_1 = 0.9, \beta_2 = 0.999)$ and a constant learning rate of $1e{-3}$. 

Other training details, hyperparameters, and the code to reproduce our results can be found at \href{https://github.com/MECLabTUDA/GAUDA/}{https://github.com/MECLabTUDA/GAUDA/}.

%% file: sections_camera_ready/05_results.tex
\section{Results}

In this section, we present the performance metrics, comparative analyses, and statistical significance of our proposed method's outcomes on surgical segmentation datasets.

\subsection{Image Quality Assessment}

\textbf{Qualitative examples} of generated (image, mask) pairs are displayed in Figure \ref{fig:qual}. The joint modelling of images and masks using an LDM allows for high-quality and semantically coherent synthesis of paired segmentation data. CFG enables class-controlled synthesis with good variety.

\begin{figure}[ht]
    \centering
    \if\usepng1
        \includegraphics[width=\linewidth]{figures/ldm_eval_single_column.png}
    \else
        \includegraphics[width=\linewidth]{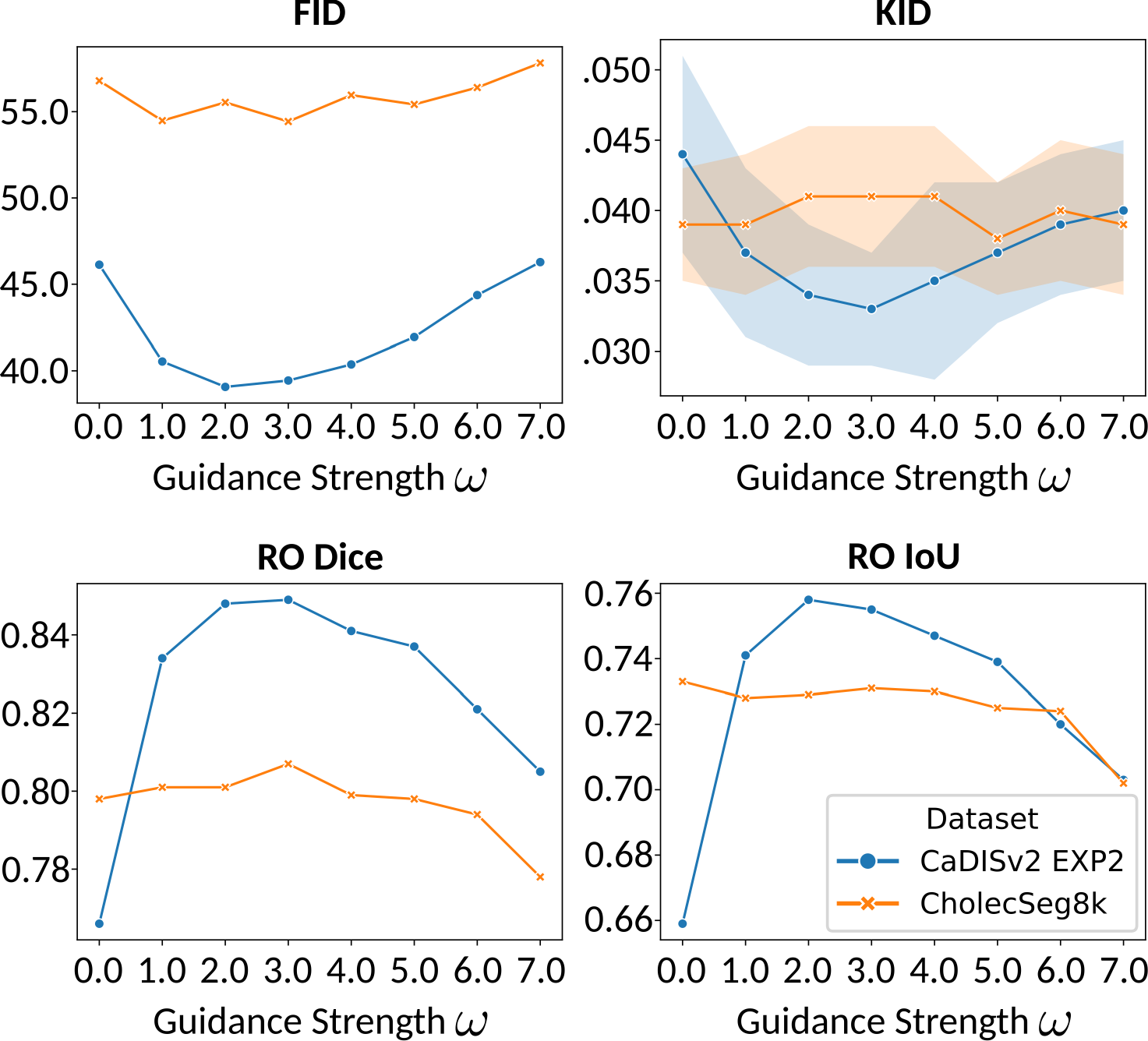}
    \fi
    \caption{\textbf{Quantitative Assessment of the Generative Model.} Our generative model achieves optimal performance with a guidance strength $\omega \in [2.0, 3.0]$. While altering $\omega$ has a medium effect on fidelity (top row), especially for CholecSeg8k, the decrease in terms of \emph{Real Only} metrics (RO, bottom row), is more severe.}
    \label{fig:quan}
\end{figure}

To assess the objective quality of synthesised examples from our generative model, we \textbf{quantitatively evaluate} our generative model using the \textbf{FID} \cite{heusel2017gans} and \textbf{KID} \cite{binkowski2018demystifying} metrics on $4096$ synthesised examples. Further, since we are interested in generating coherent (image, mask) pairs, we conduct two more analyses. Both quantify the overall semantic alignment and expressiveness of generated paired data beyond purely addressing image quality:

Firstly, we train PHiSeg on the original training data and evaluate this model's Dice and IoU scores on the $4096$ synthesised samples, denoted as \textbf{Real Only (RO)}. By doing this, we can quantify the realism and coherence of the synthesised pairs. As the results displayed in Figure \ref{fig:quan} show, our generative model is capable of synthesising high-quality (image, mask) pairs with great semantic alignment, optimal with a guidance strength $\omega \in [2.0,3.0]$. Quantitative comparisons against DatasetGAN \cite{zhang2021datasetgan} are presented in Section \ref{supp:gan} in the Supplementary. Section \ref{supp:efficiency} further gives an overview of training times and inference speed of each method, component and training scheme.

Secondly, we train the same method solely on the $4096$ synthetic pairs ($\omega=3.0$) until convergence and evaluate the IoU score on the original test data. Interestingly, PHiSeg \textbf{Synthetic Only} performs already very close to PHiSeg trained on the real training data ($0.635$ vs. $0.738$ IoU for CaDISv2 and $0.742$ vs. $0.854$ IoU for CholecSeg8k, averaged over labels). This demonstrates the expressiveness, diversity and coherence of the generated pairs.

As a boundary of our generative approach, we observed that on rare occasions, the LDM generates a latent that can look noisy in image space or decode into a somewhat incoherent segmentation mask, as described in Supplementary Section \ref{supp:fail}. Here, an additional curation or error propagation reduction mechanism for generated data could further enhance the performance \cite{van2023synthetic}. We will address this point in future research.

\begin{table}[htbp]
    \centering
    \caption{\textbf{PHiSeg ($\Phi$) CholecSeg8k} Testsplit IoU per Label}
    \begin{adjustbox}{max width=0.9\linewidth}{%
        \begin{tabular}{l|cccc|cc}
            \hline
            \bfseries Augmentation (aug) & & \ding{51} & & \ding{51} & & \ding{51}\\
            \bfseries Adaptive Sampling (AS) & & & \ding{51} & \ding{51} & & \\
            \bfseries GAUDA & & & & & \ding{51} & \ding{51}\\
            \hline
            Abdominal Wall & 0.935 & 0.917 & 0.942 & 0.923 & \textbf{0.962} & 0.945\\
            Black Background & 0.964 & 0.959 & 0.974 & 0.964 & \textbf{0.994} & 0.987\\
            Blood & 0.678 & 0.586 & \textbf{0.718} & 0.623 & 0.707 & 0.597\\
            Connective Tissue & 0.883 & 0.842 & 0.899 & 0.856 & \textbf{0.909} & 0.872\\
            Cystic Duct & 0.801 & 0.751 & \textbf{0.817} & 0.766 & \textbf{0.819} & 0.756\\
            Fat & 0.919 & 0.905 & 0.929 & 0.910 & \textbf{0.948} & 0.933\\
            Gallbladder & 0.883 & 0.850 & 0.887 & 0.854 & \textbf{0.908} & 0.876\\
            Gastrointestinal Tract & 0.833 & 0.799 & 0.832 & 0.774 & \textbf{0.862} & 0.804\\
            Grasper & 0.842 & 0.812 & 0.847 & 0.804 & \textbf{0.865} & 0.823\\
            Hepatic Vein & 0.613 & 0.229 & \textbf{0.624} & 0.385 & 0.503 & 0.316\\
            L-Hook Electrocautery & 0.883 & 0.851 & 0.890 & 0.856 & \textbf{0.907} & 0.864\\
            Liver & 0.925 & 0.909 & 0.932 & 0.912 & \textbf{0.953} & 0.937\\
            Liver Ligament & 0.945 & 0.934 & 0.953 & 0.946 & \textbf{0.973} & 0.964\\
            \hline
            Label Mean & 0.854 & 0.796 & 0.865 & 0.813 & \textbf{0.870} & 0.821\\
            \hline
            Sample Mean & 0.899 & 0.875 & 0.907 & 0.878 & \textbf{0.926} & 0.898\\
            \hline
            Sample Median & 0.925 & 0.907 & 0.933 & 0.913 & \textbf{0.952} & 0.935 
        \end{tabular}
    }
    \end{adjustbox}%
    \label{tab:cholecseg_phi}
\end{table}

\subsection{Segmentation Downstream Task}

\begin{figure*}[htbp]
    \centering
    \if\usepng1
        \includegraphics[width=0.83\textwidth]{figures/cholecseg_iou_violinplots.png}
    \else
        \includegraphics[width=0.83\textwidth]{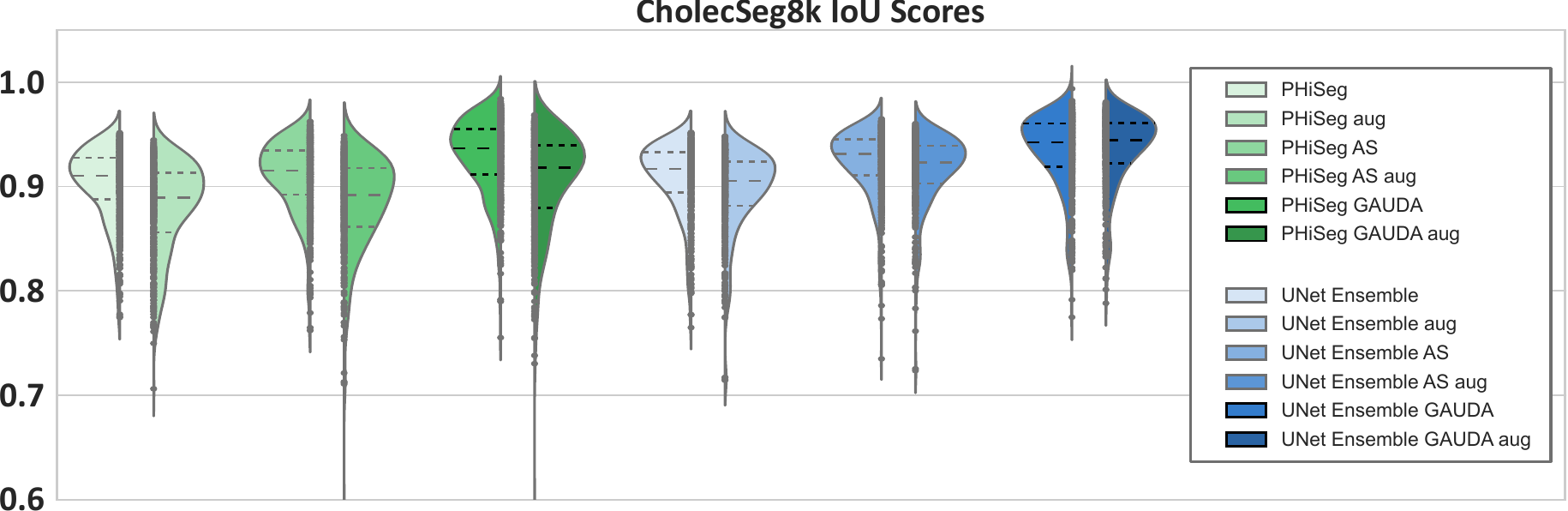}
    \fi
    \caption{\textbf{CholecSeg8k Downstream IoU Results.} The vertical axis represents IoU values in $[0,1]$ range. The horizontal axis corresponds to different base models (PHISeg and UNet Ensemble) and training types (No augmentation/oversampling, classic augmentation (\emph{aug}), Adaptive Sampling (\emph{AS}) and \emph{GAUDA}). DICE and AP scores can be found in the Supplementary.}
    \label{fig:downstream_cholec}
\end{figure*}

To highlight the performance improvements from synthetic (image, mask) pairs, we use \emph{GAUDA} while training PHiSeg ($\Phi$) and the UNet ensemble ($\Upsilon$) on both datasets. We compare this approach to training the models with Adaptive Sampling (\emph{AS}) and training without resampling. Further, we evaluate every method trained with classic data augmentation (\emph{aug}), consisting of cropping away 10\% of the image, horizontal flipping, and rotation up to 45°, each applied with a probability of 20\%. \emph{GAUDA} outperforms comparable oversampling and augmentation methods for surgical segmentation by an absolute IoU of 1.6\% on CaDISv2 Setting II and 1.5\% on CholecSeg8k. The sample-wise and average IoU scores are visualised in Figure \ref{fig:downstream_cholec} for CholecSeg8k and Figure \ref{fig:downstream_cadis} for CaDISv2. Additional DICE and Average Precision (AP) scores can be found in Supplementary Figure \ref{fig:downstream_dice_ap}. Detailed PHiSeg IoU scores per label are further provided in Table \ref{tab:cholecseg_phi} for CholecSeg8k and Table \ref{tab:cadis_phi} for CaDISv2. The scores for the UNet Ensemble are given in Table \ref{tab:cholecseg_u} for CholecSeg8k and Table \ref{tab:cadis_u} for CaDISv2.

\begin{table}[htbp]
    \centering
    \caption{\textbf{UNet Ensemble ($\Upsilon$) CholecSeg8k} Testsplit Label IoU}
    \begin{adjustbox}{max width=0.9\linewidth}{%
        \begin{tabular}{l|cccc|cc}
            \hline
            \bfseries Augmentation (aug) & & \ding{51} & & \ding{51} & & \ding{51}\\
            \bfseries Adaptive Sampling (AS) & & & \ding{51} & \ding{51} & & \\
            \bfseries GAUDA & & & & & \ding{51} & \ding{51}\\
            \hline
            Abdominal Wall & 0.939 & 0.931 & 0.949 & 0.947 & \textbf{0.966} & \textbf{0.965}\\
            Black Background & 0.964 & 0.961 & 0.973 & 0.973 & \textbf{0.993} & \textbf{0.993}\\
            Blood & 0.713 & 0.685 & \textbf{0.743} & 0.718 & 0.718 & 0.727\\
            Connective Tissue & 0.894 & 0.880 & 0.909 & 0.902 & \textbf{0.918} & \textbf{0.920}\\
            Cystic Duct & 0.818 & 0.797 & \textbf{0.846} & 0.841 & 0.838 & 0.836\\
            Fat & 0.928 & 0.922 & 0.941 & 0.936 & \textbf{0.957} & \textbf{0.958}\\
            Gallbladder & 0.891 & 0.877 & 0.908 & 0.897 & 0.915 & \textbf{0.920}\\
            Gastrointestinal Tract & 0.850 & 0.826 & 0.869 & 0.855 & \textbf{0.878} & \textbf{0.879}\\
            Grasper & 0.852 & 0.821 & 0.872 & 0.857 & 0.871 & \textbf{0.876}\\
            Hepatic Vein & 0.325 & 0.232 & \textbf{0.463} & 0.298 & 0.287 & 0.293\\
            L-Hook Electrocautery & 0.884 & 0.863 & 0.898 & 0.890 & 0.890 & \textbf{0.906}\\
            Liver & 0.932 & 0.924 & 0.932 & 0.940 & \textbf{0.959} & \textbf{0.960}\\
            Liver Ligament & 0.950 & 0.948 & 0.961 & 0.959 & \textbf{0.977} & \textbf{0.979}\\
            \hline
            Label Mean & 0.842 & 0.821 & \textbf{0.867} & 0.847 & 0.860 & 0.862\\
            \hline
            Sample Mean & 0.905 & 0.892 & 0.920 & 0.912 & \textbf{0.930} & \textbf{0.932}\\
            \hline
            Sample Median & 0.931 & 0.923 & 0.944 & 0.939 & \textbf{0.958} & \textbf{0.959} 
        \end{tabular}
    }
    \end{adjustbox}%
    \label{tab:cholecseg_u}
\end{table}
As a point for further improvement of our approach, we identify the training time increase from the online inclusion of DDM-based augmentation. This is due to the slow inference speed of this model family. Currently, our method uses DDPM sampling \cite{ho2020denoising} with 1000 denoising steps. This could seamlessly be replaced by faster sampling variants such as DDIM \cite{song2020denoising} or distillation-based methods such as Consistency Models \cite{song2023consistency}. However, all these approaches introduce a trade-off between fidelity and sampling speed.

In summary, with a Cohen's d of $0.714$ (averaged across datasets and baseline models), our approach shows a medium to large effect size for improving prediction scores over comparable augmentation and resampling methods.

\begin{figure*}[htbp]
    \centering
    \if\usepng1
        \includegraphics[width=0.83\textwidth]{figures/cadis_iou_violinplots.png}
    \else
        \includegraphics[width=0.83\textwidth]{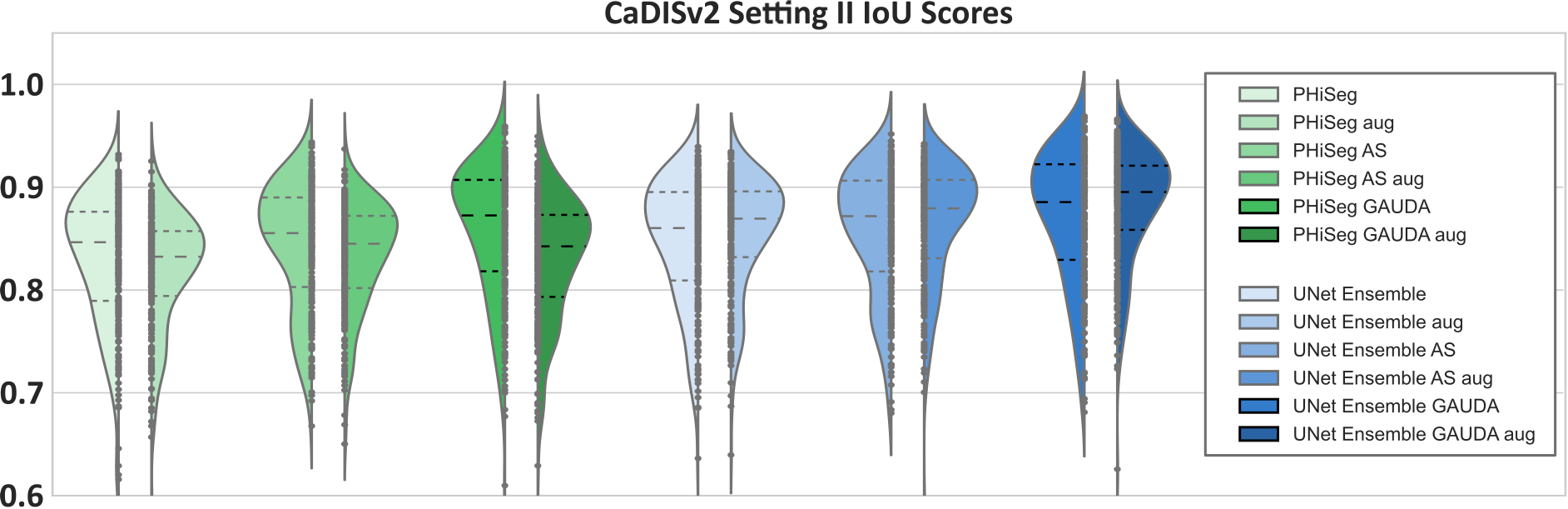}
    \fi
    \caption{\textbf{CaDISv2 Setting II Downstream IoU Results.} The vertical axis represents test-set metric values in $[0,1]$ range. The horizontal axis corresponds to different base models (PHISeg and UNet Ensemble) and training types (No augmentation/oversampling, classic augmentation (\emph{aug}), Adaptive Sampling (\emph{AS}) and \emph{GAUDA}). DICE and AP scores can be found in the Supplementary.}
    \label{fig:downstream_cadis}
\end{figure*}

%% file: sections_camera_ready/06_discussion.tex
\section{Discussion}

As shown in Tables \ref{tab:cholecseg_phi}, \ref{tab:cholecseg_u}, \ref{tab:cadis_phi} and \ref{tab:cadis_u}, \emph{GAUDA} can significantly improve performance across a variety of domains for surgical segmentation and different types of downstream models (PHiSeg and UNet Ensemble). 

\begin{figure}[H]
    \centering
    \if\usepng1
        \includegraphics[width=\linewidth]{figures/aug_no_aug_cases_v2.png}
    \else
        \includegraphics[width=\linewidth]{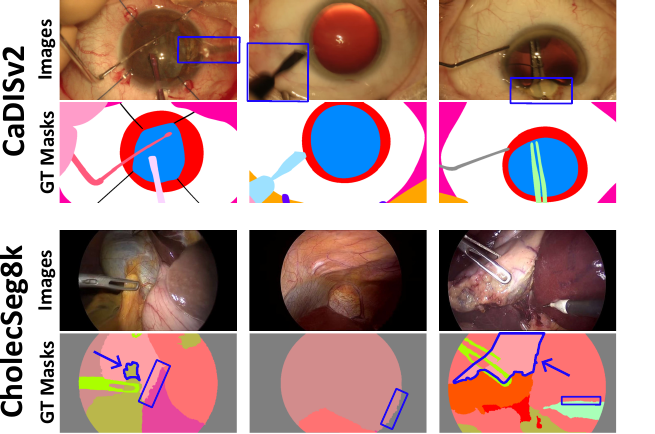}
    \fi
    \caption{\textbf{Samples with Significant Performance Changes from Augmentation.} Classic data augmentation (\emph{aug}) has a significant impact on CaDISv2 samples that are artifacted from blur or liquids (top row, blue markers). A similar effect is observable for CholecSeg8k samples with imperfect ground truth segmentation masks (bottom row, blue markers). Both are best viewed by zooming in on the digital version.}
    \label{fig:aug_no_aug}
\end{figure}

UNet Ensemble ($\Upsilon$) with the training variant of either \emph{GAUDA} or \emph{GAUDA + aug} achieves the best performance for 10 / 13 of the semantic labels of CholecSeg8k and 15 / 17 classes of CaDISv2. Similarly, PHiSeg ($\Phi$) trained with \emph{GAUDA} or \emph{GAUDA + aug} outperforms other training schemes for  11 / 13 classes of CholecSeg8k and 11 / 17 classes of CaDISv2. 

\begin{table}[htbp]
    \centering
    \caption{\textbf{PHiSeg ($\Phi$) CaDISv2} Testsplit IoU per Label}
    \begin{adjustbox}{max width=0.9\linewidth}{%
        \begin{tabular}{l|cccc|cc}
            \hline
            \bfseries Augmentation (aug) & & \ding{51} & & \ding{51} & & \ding{51}\\
            \bfseries Adaptive Sampling (AS) & & & \ding{51} & \ding{51} & & \\
            \bfseries GAUDA & & & & & \ding{51} & \ding{51}\\
            \hline
            Cannula & 0.593 & 0.560 & 0.596 & \textbf{0.613} & \textbf{0.612} & 0.543 \\
            Capsulorhexis Cystotome & 0.572 & 0.620 & 0.549 & \textbf{0.648} & 0.569 & 0.555 \\
            Capsulorhexis Forceps & 0.473 & 0.585 & 0.589 & \textbf{0.689} & 0.641 & 0.413 \\
            Cornea & 0.932 & 0.925 & 0.942 & 0.935 & \textbf{0.962} & 0.953 \\
            Eye Retractors & 0.768 & 0.704 & 0.787 & 0.729 & \textbf{0.784} & 0.690 \\
            Hand & 0.756 & 0.802 & 0.803 & 0.807 & \textbf{0.830} & 0.800 \\
            Irrigation/Aspiration Handpiece & 0.673 &	0.729 & 0.683 & 0.752 &	0.690 & \textbf{0.761} \\
            Iris & 0.829 & 0.812 & 0.844 & 0.823 & \textbf{0.857} & 0.833 \\
            Lens Injector & 0.795 & 0.788 &	0.812 & 0.786 &	\textbf{0.824} & 0.794 \\
            Micromanipulator & 0.489 & 0.541 & 0.506 & \textbf{0.559} & 0.515 & 0.475 \\
            Phacoemulsification Handpiece & 0.740 & 0.749 & 0.716 & \textbf{0.767} & 0.761 & 0.709 \\
            Primary Knife & 0.752 & 0.839 & 0.770 & \textbf{0.844} & 0.795 & 0.841 \\
            Pupil & 0.914 & 0.909 & 0.925 & 0.921 & \textbf{0.943} & 0.931 \\
            Secondary Knife & 0.775 & 0.805 & 0.823 & 0.815 & \textbf{0.852} & 0.781 \\
            Skin & 0.870 & 0.854 & 0.882 & 0.862 & \textbf{0.897} & 0.875 \\
            Surgical Tape & 0.878 & 0.851 & 0.898 & 0.863 & \textbf{0.902} & 0.878 \\
            Tissue Forceps & 0.739 & 0.709 & 0.758 & 0.758 & \textbf{0.765} & 0.734 \\
            \hline
            Label Mean & 0.738 & 0.753 & 0.758 & 0.775 & \textbf{0.786} & 0.739 \\
            \hline
            Sample Mean & 0.829 & 0.817 & 0.843 & 0.832 & \textbf{0.857} & 0.828 \\
            \hline
            Sample Median & 0.886 & 0.867 & 0.897 & 0.875 & \textbf{0.910} & 0.884 
        \end{tabular}
    }%
    \end{adjustbox}
    \label{tab:cadis_phi}
\end{table}

\begin{table}[htbp]
    \centering
    \caption{\textbf{UNet Ensemble ($\Upsilon$) CaDISv2} Testsplit IoU per Label}
    \begin{adjustbox}{max width=0.9\linewidth}{%
        \begin{tabular}{l|cccc|cc}
            \hline
            \bfseries Augmentation (aug) & & \ding{51} & & \ding{51} & & \ding{51}\\
            \bfseries Adaptive Sampling (AS) & & & \ding{51} & \ding{51} & & \\
            \bfseries GAUDA & & & & & \ding{51} & \ding{51}\\
            \hline
            Cannula & 0.617 & 0.690 & 0.635 & 0.699 & 0.639 & \textbf{0.702} \\
            Capsulorhexis Cystotome & 0.597 & 0.677 & 0.601 & 0.677 & 0.611 & \textbf{0.705} \\
            Capsulorhexis Forceps & 0.539 & 0.711 & 0.545 & \textbf{0.756} & 0.478 & 0.674 \\
            Cornea & 0.936 & 0.939 & 0.946 & 0.948 & \textbf{0.965} & \textbf{0.967} \\
            Eye Retractors & 0.798 & 0.773 & 0.815 & 0.781 & \textbf{0.827} & 0.801 \\
            Hand & 0.849 & 0.862 & 0.858 & 0.885 & 0.884 & \textbf{0.911} \\
            Irrigation/Aspiration Handpiece & 0.727 & 0.761 & 0.734 & 0.767 & 0.741 & \textbf{0.778} \\
            Iris & 0.844 & 0.851 & 0.853 & 0.861 & 0.868 & \textbf{0.879} \\
            Lens Injector & 0.791 & \textbf{0.821} & 0.792 & 0.802 & 0.800 & 0.816 \\
            Micromanipulator & 0.494 & 0.542 & 0.509 & 0.582 & 0.501 & \textbf{0.589} \\
            Phacoemulsification Handpiece & 0.743 & 0.765 & 0.744 & 0.774 & 0.749 & \textbf{0.818} \\
            Primary Knife & 0.843 & 0.882 & 0.855 & 0.891 & 0.889 & \textbf{0.913} \\
            Pupil & 0.920 & 0.926 & 0.930 & 0.936 & 0.947 & \textbf{0.955} \\
            Secondary Knife & 0.865 & 0.861 & 0.855 & 0.876 & \textbf{0.889} & 0.882 \\
            Skin & 0.878 & 0.882 & 0.887 & 0.894 & 0.905 & \textbf{0.908} \\
            Surgical Tape & 0.887 & 0.883 & 0.891 & 0.895 & 0.908 & \textbf{0.912} \\
            Tissue Forceps & 0.803 & 0.782 & 0.813 & 0.794 & \textbf{0.820} & 0.810 \\
            \hline
            Label Mean & 0.772 & 0.801 & 0.780 & 0.813 & 0.789 & \textbf{0.825} \\
            \hline
            Sample Mean & 0.846 & 0.853 & 0.855 & 0.864 & 0.870 & \textbf{0.881} \\
            \hline
            Sample Median & 0.898 & 0.903 & 0.908 & 0.912 & 0.925 & \textbf{0.928} 
        \end{tabular}
    }%
    \end{adjustbox}
    \label{tab:cadis_u}
\end{table}

The classic augmentation scheme (\emph{aug}) often decreases the models' performance, especially for PHiSeg. We accredit this to the augmentations potentially being too drastic, hindering generalisation. Especially for CaDISv2, the dataset never contains any rotated recordings of the eye, and tools typically appear from specific sides and angles. 

Only for the UNet Ensemble ($\Upsilon$), trained on CaDISv2, \emph{aug} generally improves the results further. Additionally, for CholecSeg8k, \emph{GAUDA + aug} further improves performance over \emph{GAUDA} for the UNet Ensemble. Figures \ref{fig:downstream_cholec} and \ref{fig:downstream_cadis} reveal that performance especially improves for the long tail of the data distribution with lower initial performance. 

Another important observation is the inconsistent impact of \emph{aug} on the performance of downstream task models. This becomes visible in the scores for \emph{GAUDA} versus \emph{GAUDA aug}, highlighted with darker lines in Figures \ref{fig:downstream_cholec} and \ref{fig:downstream_cadis}. The classic augmentation scheme aids the UNet Ensemble but harms PHiSeg. To investigate, we visually inspect samples for which we obtain the most significant IoU increase of UNet Ensemble and decrease for PHiSeg when comparing \emph{GAUDA} and \emph{GAUDA aug}. The three test samples of each dataset with the largest gap are displayed in Figure \ref{fig:aug_no_aug}. For CaDISv2, these samples show a lot of artifacts due to blurring and/or liquids. Meanwhile, the CholecSeg8k examples yield noisy, imperfect ground truth segmentation masks. Understanding the impact of data augmentation - classic or generative - is an important topic for future research. 

%% file: sections_camera_ready/07_conclusions.tex
\section{Conclusions}

We present GAUDA, a novel training method for surgical segmentation, which leverages the epistemic uncertainty of a Bayesian downstream task model to guide online augmentation by generative models. We further demonstrate how LDMs can successfully be used for the generative augmentation of surgical segmentation data. By learning a joint representation of latent image and mask encodings, we can synthesise unseen (image, mask) pairs with great quality and outstanding semantic coherence. Therefore, our approach has the potential to drastically decrease costs and effort for data allocation and annotation in the surgical domain. Utilising such synthetic data in our GAUDA approach, we can boost the performance of Bayesian segmentation models by a significant amount without having to gather additional training samples. As a result, we hope that our approach can contribute to raising methods for analysing surgical scenes, and for automated surgical assistance, to a new unmatched level of performance.

\paragraph{Acknowledgements}

This work has been partially funded by the \textit{German Federal Ministry of Education and Research} as part of the \textit{Software Campus} programme (research grant 01$|$S23067). 

%% file: sections_camera_ready/supplementary.tex
\section{GAUDA - Supplementary}

This chapter provides additional supplementary information for \emph{GAUDA: \textit{Generative Adaptive Uncertainty-guided Diffusion-based Augmentation} for Surgical Segmentation}.

\subsection{GAUDA Pseudocode}
\label{supp:pseudo}

\begin{figure}[ht]
    \centering
	\includegraphics[width=\linewidth]{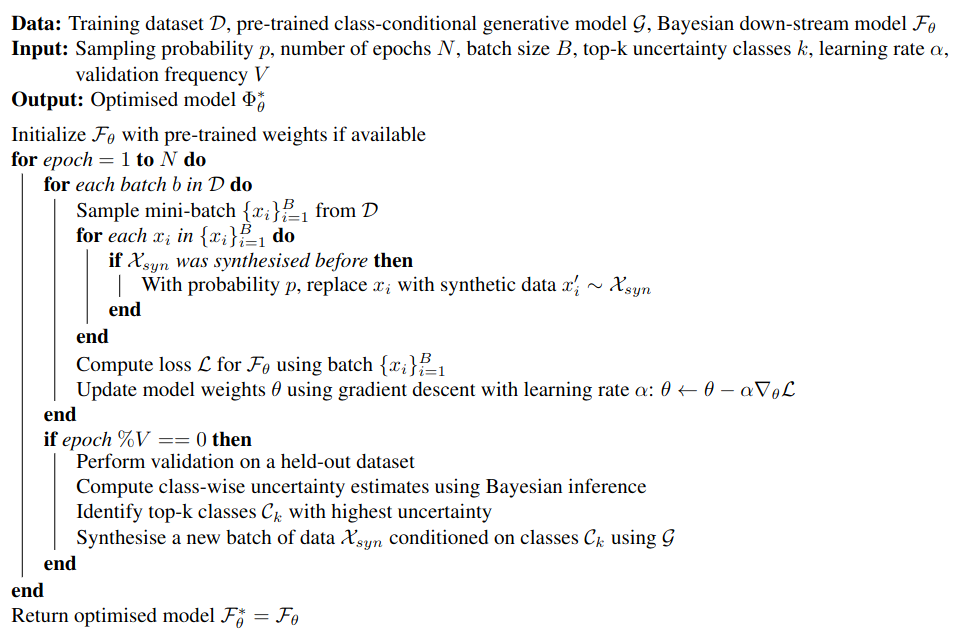}
    \caption{\textbf{GAUDA Pseudocode.}}
    \label{fig:pseudocode}
\end{figure}

A pseudocode formulation for training Bayesian downstream task models with GAUDA can be found in Figure \ref{fig:pseudocode}. GAUDA serves as a training scheme with flexible options for the downstream task, downstream model and uncertainty estimation method.

\subsection{Uncertainty-based Sampling Versus Score-based Sampling}
\label{supp:sampling}

\begin{figure}[ht]
    \centering
    \if\usepng1
        \includegraphics[width=0.92\linewidth]{figures/toy_example.png}
    \else
        \includegraphics[width=0.92\linewidth]{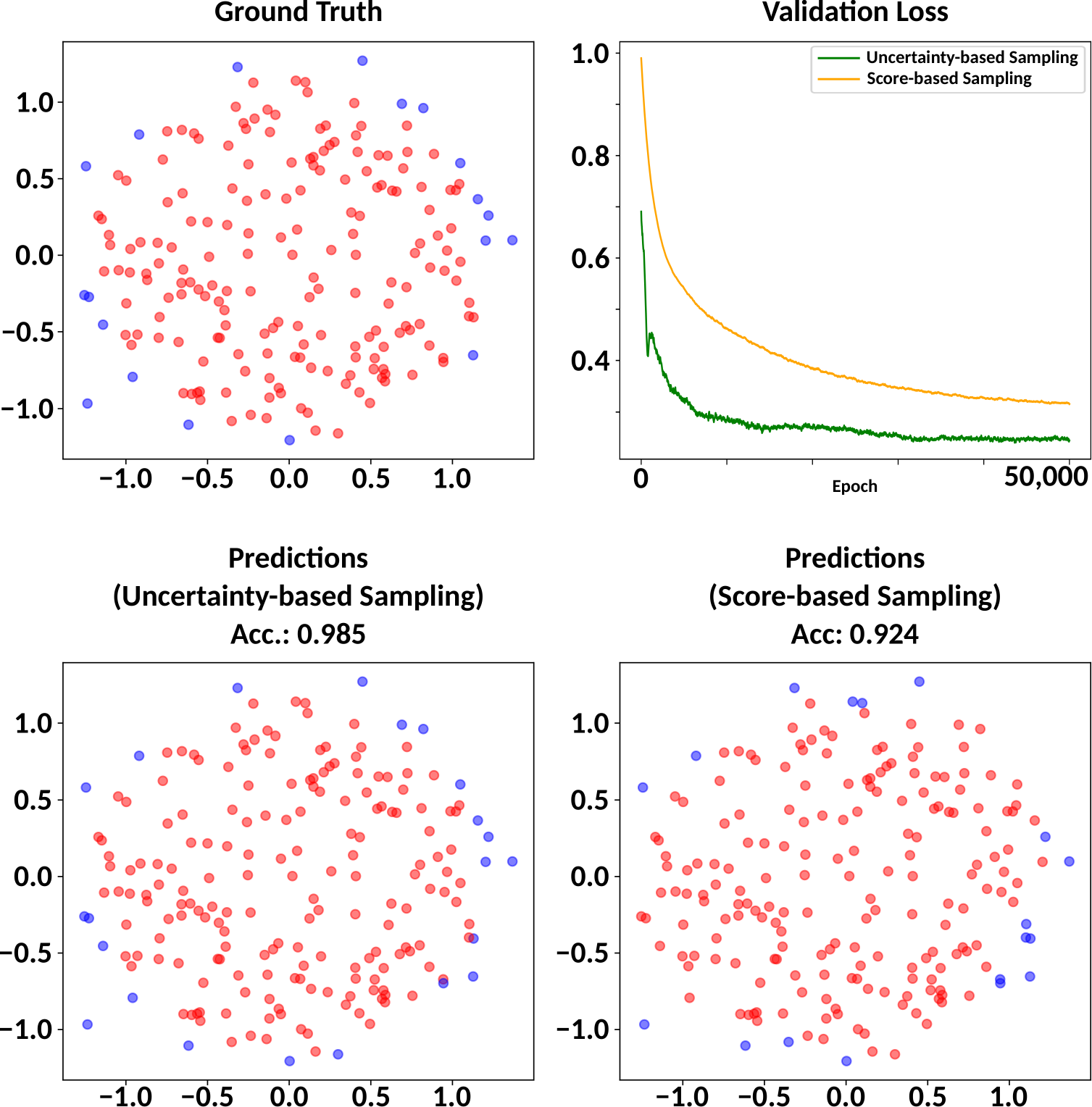}
    \fi
    \caption{\textbf{Uncertainty- Versus Score-based Sampling.}}
    \label{fig:toy_class}
\end{figure}

In this section, we analyse the effect of the predictive epistemic uncertainty as quantity for re-defining sampling weights in adaptive sampling. For that purpose, we define a simplified classification problem of two-dimensional points. As visualised in the top left plot of Figure \ref{fig:toy_class}, the data consists of two noisy classes depending on their centre distance (red and blue). The data shows a significant imbalance in the number of samples per class. 

We deploy a simple neural network classifier with two fully connected layers, an intermediate feature size of 10 nodes and a ReLU activation + dropout with $50 \%$ chance. To obtain UE, we build an ensemble of 20 of such models.

We compare two training schemes to investigate the effect of different quantities for redefining sample weights. First, we use the validation accuracy analogously to the original AS formulation. Second, we use the epistemic uncertainty from the variance of the ensemble prediction.

As visualised in the top right and bottom plots of Figure \ref{fig:toy_class}, sampling based on UE yields faster convergence and improved generalisation capabilities, ultimately resulting in a testing accuracy improved by $6.1\%$.

\subsection{Pre-Train Augmentation Versus Online Augmentation}
\label{supp:augmentation}

Adaptive augmentation of training data leads to a change in the data distribution, favouring underrepresented data points more and more during training. To demonstrate this, we first sample a fixed amount of additional samples of our simplified experimental data (doubling the number of examples). Second, we train the deep ensemble from Section \ref{supp:sampling} with the GAUDA scheme, adaptively augmenting the data based on the predictive epistemic uncertainty (again doubling the number of examples).

\begin{figure}[ht]
    \centering
    \if\usepng1
        \includegraphics[width=0.92\linewidth]{figures/toy_aug.png}
    \else
        \includegraphics[width=0.92\linewidth]{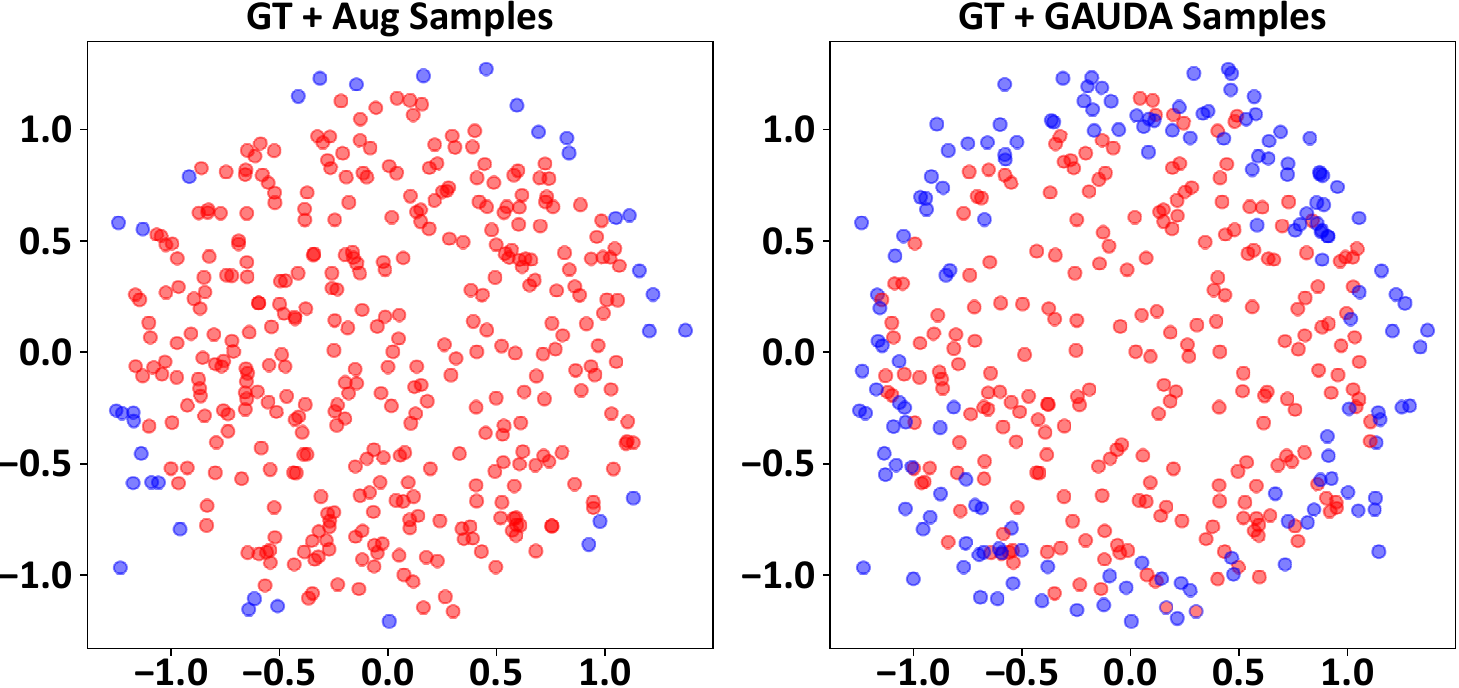}
    \fi
    \caption{\textbf{Pre-train Versus Online Augmentation.}}
    \label{fig:toy_aug}
\end{figure}

Figure \ref{fig:toy_aug} shows adaptive online augmentation yields a significantly higher percentage of samples of the limited blue class compared to pre-train random augmentation.

\subsection{Additional Segmentation Scores}
\label{supp:metrics}

\begin{figure*}[htbp]
    \centering
    \if\usepng1
        \includegraphics[width=0.9\textwidth]{figures/downstream_dice_ap_violinplots.png}
    \else
        \includegraphics[width=0.9\textwidth]{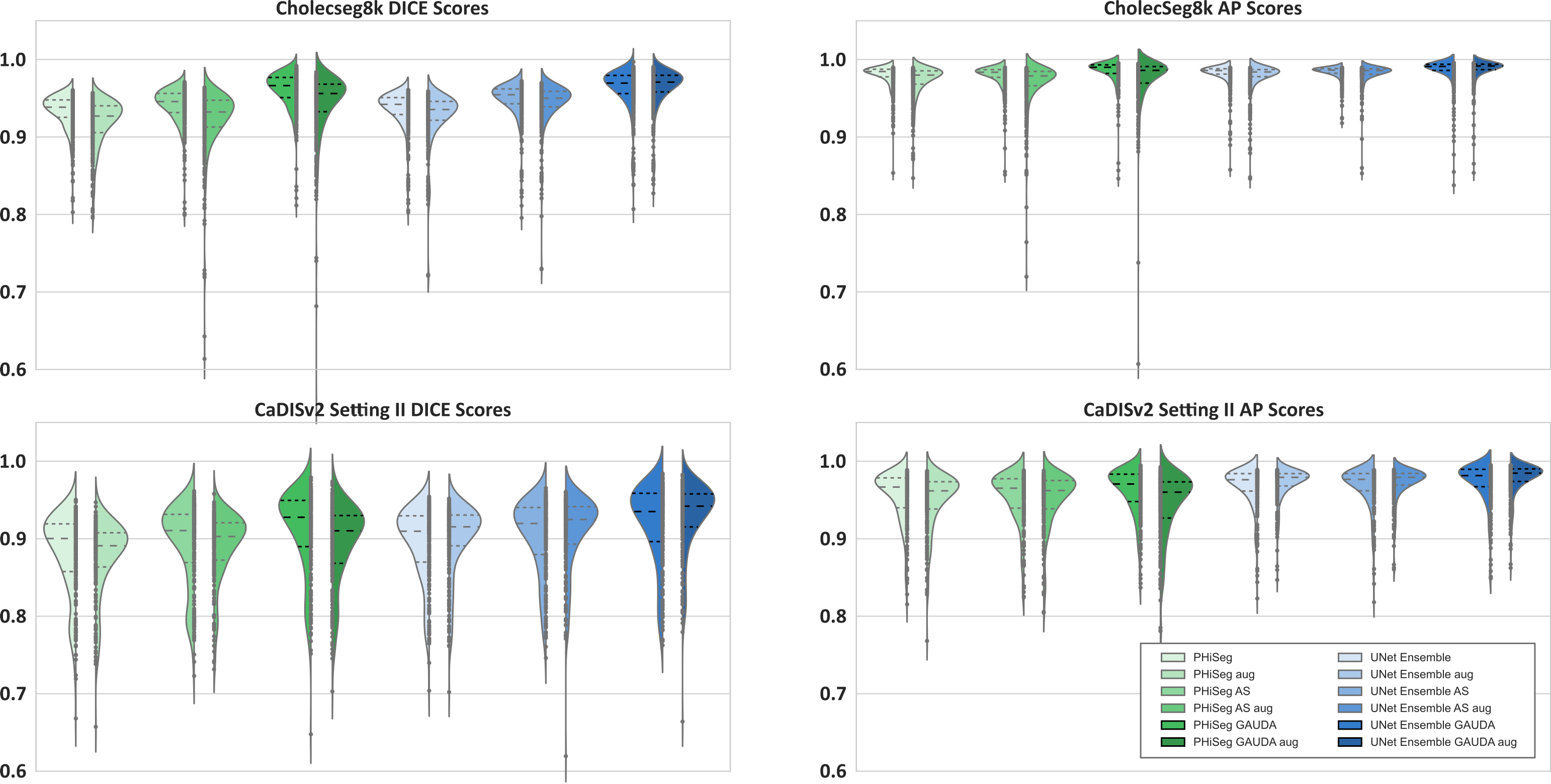}
    \fi
    \caption{\textbf{Downstream DICE and AP Results.}}
    \label{fig:downstream_dice_ap}
\end{figure*}

The sample-wise and mean DICE and Average Precision (AP) scores for the surgical segmentation downstream task are visualised in Figure \ref{fig:downstream_dice_ap}.

\subsection{Failure Cases}
\label{supp:fail}

\begin{figure}[H]
    \centering
    \if\usepng1
        \includegraphics[width=0.85\linewidth]{figures/failure_cases.png}
    \else
        \includegraphics[width=0.85\linewidth]{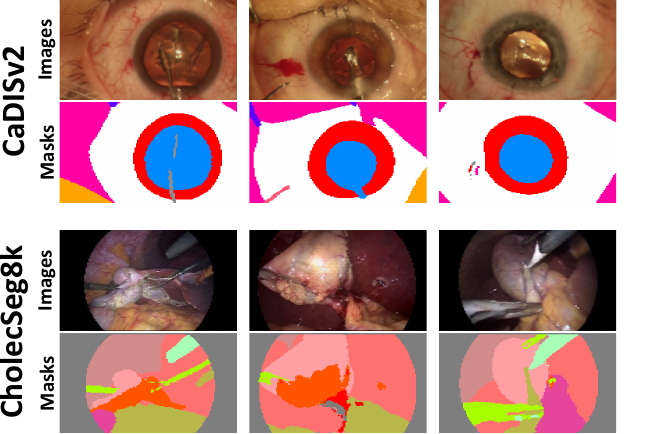}
    \fi
    \caption{\textbf{Erroneous Synthetic Samples.}}
    \label{fig:failure}
\end{figure}

Figure \ref{fig:failure} displays examples of synthetic (image, mask) pairs with improvable quality. Notably, failures can occur in the form of noisy tool segmentation masks (left column), wrongly allocated or missing labels (middle column) and small inconsistent regions (right column). Yet, erroneous regions are small and can potentially be filtered out or improved with error propagation reduction mechanisms \cite{van2023synthetic}.

\subsection{Comparison to GAN-based Approaches}
\label{supp:gan}

In Table \ref{tab:gan}, we compare the quantitative performance of our LDM ($\omega=3.0$) to DatasetGAN \cite{zhang2021datasetgan}. The DatasetGAN implementation is based on an improved version from EditGAN \cite{ling2021editgan}. It uses StyleGAN2 \cite{karras2019style} with 'config-f', trained on $2e6$ random examples of both datasets, which was sufficient for convergence. For training the interpreter model, we used 50 annotated (image, mask) pairs and early stopping based on the loss progress.
\begin{table}[ht]
    \centering
    \resizebox{0.95\columnwidth}{!}{%
        \begin{tabular}{c|c|cccc}
             \hline
             \textbf{Method} & \textbf{Dataset} & \textbf{FID} ($\downarrow$) & \textbf{KID} ($\downarrow$) & \textbf{RO IoU} ($\uparrow$) & \textbf{SO IoU} ($\uparrow$) \\
             \hline
             LDM (ours) & CaDISv2 & 39.44 & 0.033 $\pm$ 0.004 & 0.755 & 0.635 \\
             DatasetGAN \cite{zhang2021datasetgan} & CaDISv2 & 66.99 & 0.057 $\pm$ 0.009 & 0.281 & 0.213 \\
             \hline
             LDM (ours) & CholecSeg8k & 56.80 & 0.041 $\pm$ 0.005 & 0.731 & 0.742 \\
             DatasetGAN \cite{zhang2021datasetgan} & CholecSeg8k & 65.40 & 0.042 $\pm$ 0.007 & 0.114 & 0.145
        \end{tabular}
    }
    \caption{\textbf{Quantiative Comparison against DatasetGAN.}}
    \label{tab:gan}
\end{table}

Notably, our generative model surpasses DatasetGAN in terms of fidelity, but especially in RO and SO scores, indicating a superior semantic alignment between images and masks of generated pairs.

\subsection{Computational and Resource Efficiency}
\label{supp:efficiency}

Table \ref{tab:efficiency} lists the number of training examples, the total training time and inference speed of each component of our proposed method, each component of DatasetGAN \cite{zhang2021datasetgan}, as well as each training scheme. The reported numbers are averaged over datasets and downstream task models. The inference speed is reported for a single sample.
\begin{table}[ht]
    \centering
    \resizebox{0.95\columnwidth}{!}{%
        \begin{tabular}{c|ccc}
            \hline
             \textbf{Method / Component} & \textbf{Num. Examples} & \textbf{Training Time} & \textbf{Inference Speed} \\
             \hline
             Image VQ-GAN & $1.2e6$ & 11.4h & 4ms \\
             Mask VQ-GAN & $1.2e6$ & 20.8h & 5ms \\
             LDM & $2.0e7$ & 125.1h & 36,891ms\\
             Full Model (ours) & - & - & 36,895ms \\
             \hline
             StyleGAN2 & $2.0e6$ & 36.9h & 290ms \\
             StyleGAN Encoder & $6.0e5$ & 7.1h & 9ms \\
             DatasetGAN & $\leq 4.0e6$ & 0.25h & 380ms \\
             \hline
             Downstream Default & $2.5e6$ & 15.3h & 8ms \\ 
             Adaptive Sampling & $2.5e6$ & 18.3h & 8ms \\
             GAUDA & $2.5e6$ & 22.6h & 8ms \\
        \end{tabular}
    }
    \caption{\textbf{Training Times and Inference Speed.}}
    \label{tab:efficiency}
\end{table}